\def\year{2022}\relax
\documentclass[letterpaper]{article} % DO NOT CHANGE THIS
\usepackage{aaai22}  % DO NOT CHANGE THIS
\usepackage{times}  % DO NOT CHANGE THIS
\usepackage{helvet}  % DO NOT CHANGE THIS
\usepackage{courier}  % DO NOT CHANGE THIS
\usepackage[hyphens]{url}  % DO NOT CHANGE THIS
\usepackage{graphicx} % DO NOT CHANGE THIS
\usepackage{natbib}  % DO NOT CHANGE THIS AND DO NOT ADD ANY OPTIONS TO IT
\usepackage{caption} % DO NOT CHANGE THIS AND DO NOT ADD ANY OPTIONS TO IT
\DeclareCaptionStyle{ruled}{labelfont=normalfont,labelsep=colon,strut=off} % DO NOT CHANGE THIS
\newcommand{\etal}{{\emph{et al.}}}
\usepackage{graphicx}
\usepackage{amsmath}
\usepackage{amssymb}

\usepackage{dsfont}
\usepackage{booktabs}
\usepackage{multirow}
\usepackage{color}
\usepackage[super]{nth}
\usepackage{subfigure}
\usepackage{algorithm}
\usepackage{algorithmic}

\usepackage{newfloat}
\usepackage{listings}
\floatstyle{ruled}
\newfloat{listing}{tb}{lst}{}
\floatname{listing}{Listing}

\title{Addressing Multiple Salient Object Detection via Dual-Space Long-Range Dependencies}

\title{Addressing Multiple Salient Object Detection via \\Dual-Space Long-Range Dependencies}
\author {
    Bowen Deng,
    Andrew P. French, 
    Michael P. Pound
}
\affiliations {
    Computer Vision Laboratory, School of Computer Science, University of Nottingham, Nottingham, United Kingdom\\
   bowen.deng@nottingham.ac.uk, andrew.p.french@nottingham.ac.uk, michael.pound@nottingham.ac.uk
}

\usepackage{bibentry}
\begin{document}

\maketitle

\begin{abstract}
 Salient object detection plays an important role in many downstream tasks. However, complex real-world scenes with varying scales and numbers of salient objects still pose a challenge. In this paper, we directly address the problem of detecting multiple salient objects across complex scenes. We propose a network architecture incorporating non-local feature information in both the spatial and channel spaces, capturing the long-range dependencies between separate objects. Traditional bottom-up and non-local features are combined with edge features within a feature fusion gate that progressively refines the salient object prediction in the decoder. We show that our approach accurately locates multiple salient regions even in complex scenarios. To demonstrate the efficacy of our approach to the multiple salient objects problem, we curate a new dataset containing only multiple salient objects. Our experiments demonstrate the proposed method presents state-of-the-art results on five widely used datasets without any pre-processing and post-processing. We obtain a further performance improvement against competing techniques on our multi-objects dataset. The dataset and source code are available at: https://github.com/EricDengbowen/DSLRDNet.
\end{abstract}

\section{Introduction}
Salient object detection (SOD) aims to highlight the most visually striking or important objects of a scene. SOD plays a significant role in computer vision pipelines, and has been widely applied to many object-level applications in various areas such as object recognition \citep{rutishauser2004bottom}, object detection \citep{ren2013region,zhang2017bridging}, image retrieval \citep{he2012mobile}, image captioning \citep{das2017human,fang2015captions}, weekly supervised semantic segmentation \citep{wang2018weakly,wei2016stc} and image cropping \citep{wang2018deep}.

\begin{figure}[t]
\begin{center}
\includegraphics[width=1.0\linewidth]{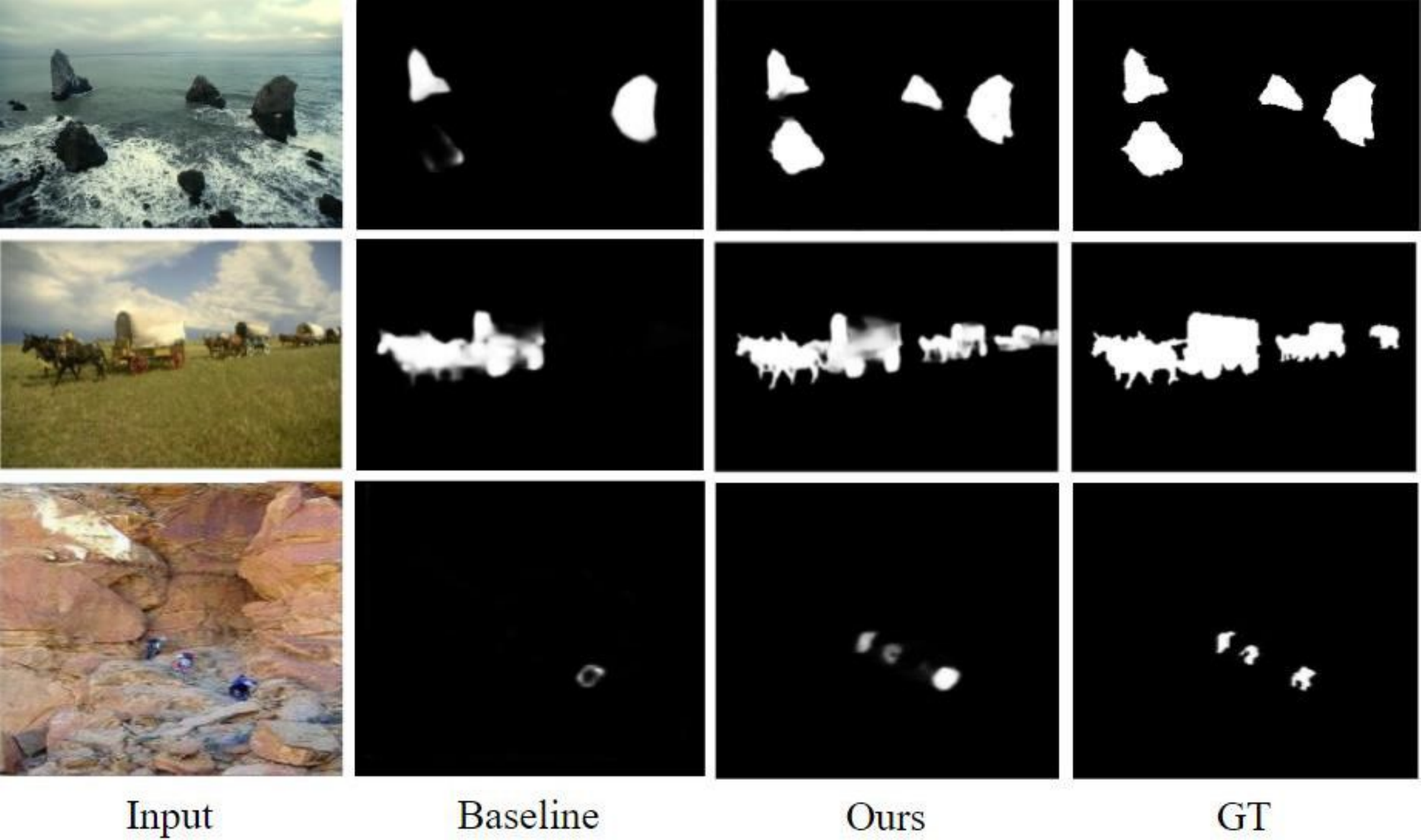}
\end{center}
   \caption{Visual examples of our proposed method. After introducing dual-space long-range dependencies, our model can handle complicated scenes with multiple salient objects.}
\label{fig:figureone}
\end{figure}

Although remarkable progress has been made, there still remain many open challenges. Existing SOD datasets contain many images with a single object, often centred in the middle of the image. Human observers may be drawn naturally to centred objects, but in complex scenes they can identify numerous salient objects distributed throughout a scene. Many existing techniques in saliency are based upon traditional U-shape networks, containing only convolutional operations that process local neighborhoods. This fails to exploit the long-range pixel-wise or channel-wise relationships among features in an image, and ultimately leaves techniques less able to address the problem of multiple salient objects (Fig~\ref{fig:figureone}). Long-range dependencies have been shown to play an important role in different classification tasks~\citep{wang2018non}, and this is also true of pixel-level segmentation works such as salient object detection. The size of the receptive field is significant in locating and segmenting salient objects across the image. Large kernels aid segmentation tasks but the experimental receptive fields are usually smaller than the ones in theory~\citep{liu2019simple, zhao2017pyramid, peng2017large}. This will likely limit the performance of SOD networks in which objects are spatially separated. 

Recent state-of-the-art SOD approaches solve the problem of salient object detection in the general case, often by combining and refining multi-level features into feature representations \citep{hou2017deeply, zhang2018bi,wu2019cascaded,wu2019stacked,zhao2019egnet,liu2019simple}, introducing additional losses into frameworks to provide structural information \citep{feng2019attentive,qin2019basnet} or applying attention mechanisms to filter the redundant information and focus on the valuable features \citep{zhang2018progressive,liu2018picanet, chen2018reverse,feng2019attentive,wu2019cascaded}. Only few of the existing SOD methods consider long-range dependencies \citep{liu2020learning,li2020multi,zhou2020multi,sun2019self}, and none of these works specifically address the problem of multiple salient objects.

In this paper, we propose a novel architecture for multiple salient object detection that considers long-range dependencies in both spatial space and channel space. Inspired by \citep{fu2019dual,wang2018non}, we propose a non-local guidance module (NLGM),  comprising several dual-space non-local blocks (DSNLBs) to capture pixel-wise and channel-wise relationships. Features at each position are aggregated by a weighted sum of all the features in spatial space, while each channel map is updated by a weighted integration of all interdependent channel maps. Unlike previous work, we stack several DSNLBs in order to capture non-local features in a progressive manner. Bottom-up convolutional features and these non-local features are combined in the network decoder through feature fusion gates, that control the transmission of information into the next stage of the network. These include salient edge supervision to further enhance the quality of the saliency maps. We demonstrate the improved performance of our network on numerous datasets, and specifically multiple salient object detection (MSOD) problems by evaluating on a dataset containing only complex multi-saliency images. Our contributions are:
\begin{itemize}
    \item We propose a novel MSOD framework that models long-range dependencies in both spatial space and channel space. To the best of our knowledge, this is the first paper that explicitly models long-range dependencies in this dual space for standard SOD and MSOD problems.
    \item We utilise non-local guidance and edge refinement modules that work complementarily to enrich feature representations at each stage of the top-down pathway. Features from all components are combined within a feature fusion gate, which utilises edge features to promote relevant salient and non-local features. The fusion gate ensures that top-down features are passed discriminatively through the network.
    \item We curate a new dataset containing only multiple salient objects, drawn from popular datasets in this field. We compare the proposed approach against 14 state-of-the-art methods on five widely used SOD benchmarks and the proposed multi-object dataset. Without any pre-processing and post-processing, our proposed method exceeds all previous state-of-the-art approaches in three evaluation metrics and provides a further performance boost against competing techniques on our proposed dataset.
\end{itemize}

\begin{figure*}
\includegraphics[width=0.95\textwidth]{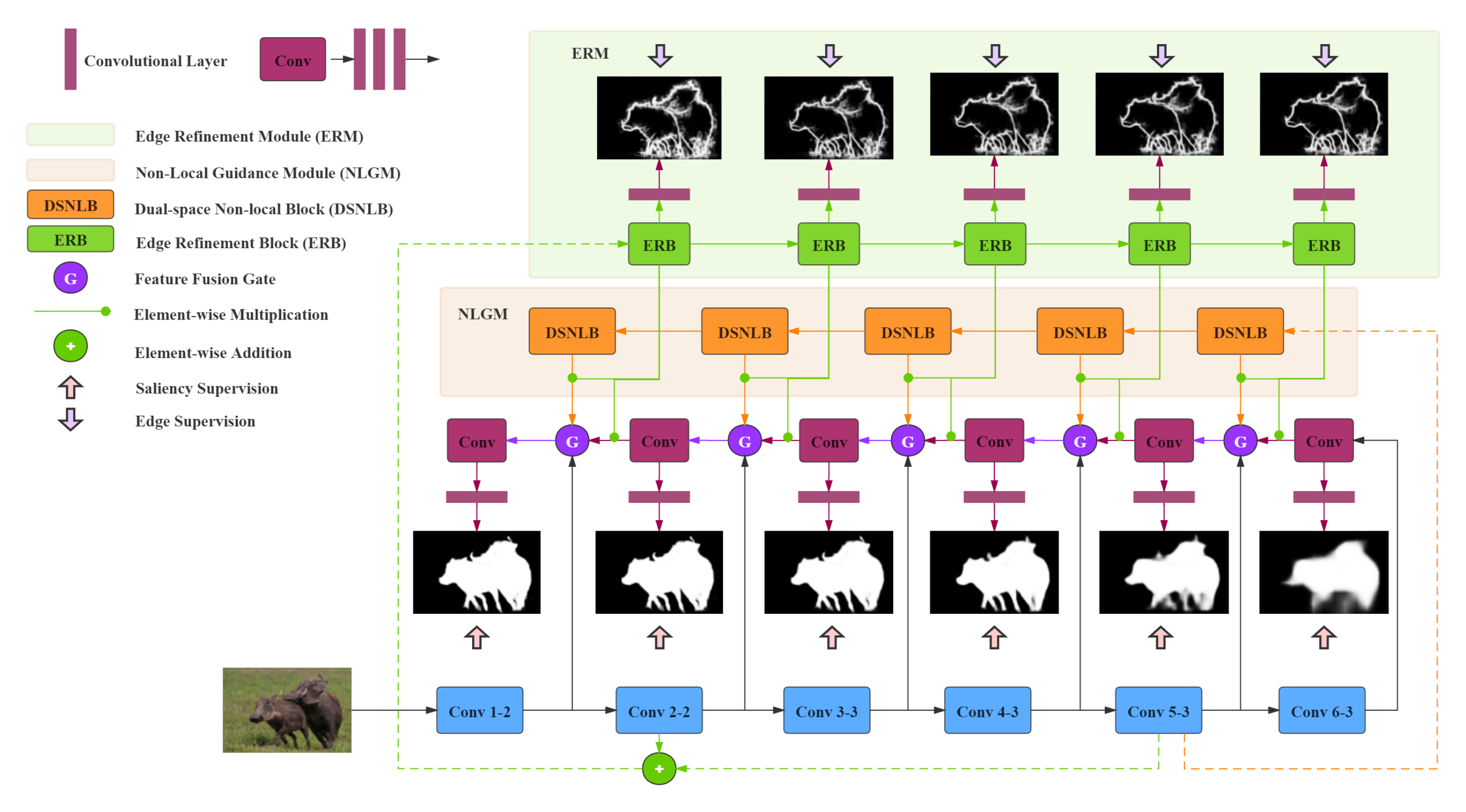}
\caption{The overall pipeline of our proposed approach, here shown using a VGG backbone. The red, orange and green boxes capture saliency features, non-local features and edge features respectively. Element-wise multiplication operates between each pair of ERB-DSNLB (edge and non-local features) and ERB-Conv (edge and saliency features). Our final prediction map is generated based on the fusion of 6 multi-scale saliency features in top-down pathway.}
\label{fig:overallpipepline}
\end{figure*}

\section{Related Work}
Traditional salient object detection methods mostly rely on low-level features \citep{zhu2014saliency,jiang2013salient} or heuristic priors such as colour contrast \citep{cheng2014global} and background \citep{wei2012geodesic,yang2013saliency}. More details are introduced in the survey by Borji \etal\citep{borji2019salient}.

Early deep SOD methods utilised multi-scale image patches \citep{li2015visual,wang2015deep}, offering impressive performance, but limited by the lack of spatial information found in small image patches. Since the introduction of fully convolutional networks (FCNs) \citep{long2015fully}, there have been many effective and efficient end-to-end SOD architectures. Among these, U-shape based architectures \citep{ronneberger2015u} see much use. Hou \citep{hou2019deeply,hou2017deeply} applied short connections from deeper to shallower layers, integrating high-level and low-level features. Zhang \etal\citep{zhang2018bi} used a gated pathway for bi-directional message passing and multi-level feature integration. Zhang \etal\citep{zhang2018progressive} embedded multi-path recurrent connections and spatial attention modules to generate saliency maps. Chen \etal\citep{chen2018reverse} designed a reverse attention block to emphasize non-object areas. Qin \etal\citep{qin2019basnet} applied another bottom-up/top-down architecture to refine the coarse saliency map generated from the prediction network. This work generates boundary-aware saliency maps through a hybrid loss approach. Feng \etal\citep{feng2019attentive} employed global perceptron modules and attentive feedback modules to detect global saliency and build encoder-decoder communication respectively. Zhao \etal\citep{zhao2019egnet} explicitly modeled edge features to guide the multi-scale features extracted from a U-shape structure and then fused multiple side-outputs into a final saliency map. Liu \etal\citep{liu2019simple} mainly investigated the role of pooling layer in U-shape structure. They proposed a global guidance module for transmitting localization information to top-down pathway and a feature aggregation module to further refine fused features. Wu \etal\citep{wu2019stacked} introduced a cross refinement unit to exchange mutual information between edge features and saliency features. Zhao \etal\citep{zhao2020suppress} designed a gated dual branch network incorporating a Fold-ASPP to better localise the salient objects of various scales. Pang \etal\citep{pang2020multi} proposed a transformation-interaction-fusion strategy to obtain efficient multi-scale features and a consistency-enhanced loss to deal with the imbalance issue between foreground and background.

Existing SOD methods seldom consider long-range dependencies: the sharing of information across spatially distant pixels, or between feature maps in channel space. Of those that do, Li \etal\citep{li2020multi} and Sun \etal\citep{sun2019self} applied self-attention mechanisms to capture spatial long-range context. Zhou \etal\citep{zhou2020multi} introduced a multi-type self-attention to capture pixel-level relationships for saliency detection in degraded images. Liu~\etal\citep{liu2020learning} designed a self-mutual attention to capture long-range contextual dependencies, this time in RGB-D. Of these works, none have made use of channel-wise dependencies as we do here and there is no ideal solution for multiple salient object detection. This has previously been hard to examine; existing public datasets contain some multi-object instances, but the frequency of these varies substantially. Here we curate a dataset specifically for this purpose, allowing us to focus on this problem.

\section{Proposed Method}
The architecture of our proposed method is shown in Fig~\ref{fig:overallpipepline}. Our model is based on a U-shape FCN combining a bottom-up pathway (backbone) and a top-down pathway. Similar to most deep SOD models, we use VGG network to illustrate our proposed structure. Following EGNet \citep{zhao2019egnet} and DSS \citep{hou2017deeply,hou2019deeply}, the last three fully connected layers are truncated and an additional side path is connected to the last pooling layer of VGG. This provides 6 outputs from bottom-up pathway representing the multi-level features captured from Conv1-2 to Conv6-3, which can be defined as a feature set \(S=\left\{S^1, S^2, S^3, S^4, S^5, S^6\right\}\).

Multi-scale saliency features are processed through the top-down pathway with a series of convolutional blocks, each containing 3 convolutional layers and ReLU activations. We refer to this saliency feature set as \(F=\left\{F^1, F^2, F^3, F^4, F^5, F^6\right\}\), where \(F^6\) is the saliency feature produced by the sixth convolutional block (rightmost Conv in Fig~\ref{fig:overallpipepline}) and so on. 

We leverage intermediate supervision \citep{lee2015deeply} at each convolutional block to improve training performance. For each saliency feature \(F^i\), a convolutional layer \(D^i_F\) is applied to produce a single-channel prediction. We use a cross-entropy loss, with the supervision here defined as:
\begin{multline}
    \mathcal{L}^i_F\left(F^i, W_{DF}^i\right)=-\sum_{j\in{Y^+}}\log Pred\left(y_j=1|F^i;W_{DF}^i\right) \\-\sum_{j\in{Y^-}}\log Pred\left(y_j=0|F^i;W_{DF}^i\right), i\in\left[1,6\right],
\end{multline}
where \(Y^+\) and \(Y^-\) denote the pixels in salient region and non-salient region respectively. \(W_{DF}^i\) denotes the parameters of convolutional layer \(D^i_F\). \(Pred\left(y_j=1|F^i;W_{DF}^i\right)\) denotes the salient prediction map.

\begin{figure}[t]
\begin{center}
\includegraphics[width=1.0\linewidth]{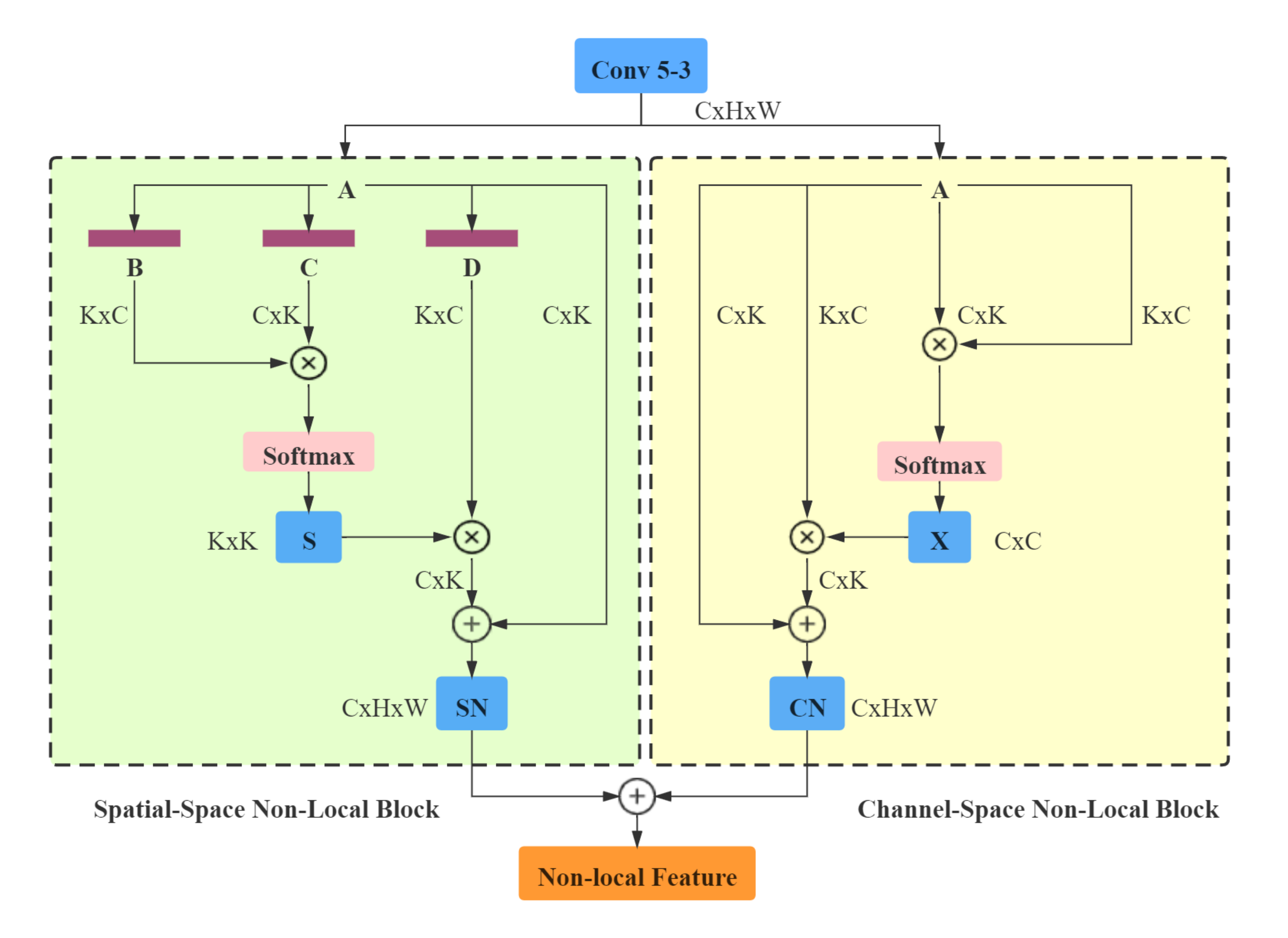}
\end{center}
   \caption{The architecture of a dual-space non-local block (DSNLB). \(C\), \(H\) and \(W\) demonstrate the channel number, height and width of given feature map respectively and \(K=H\times W\).}
\label{fig:non-local}
\end{figure}

\subsection{Non-Local Guidance Module} \label{subsubsection1}
In this module we model long-range dependencies in both spatial and channel space. Inspired by \citep{fu2019dual}, we make use of dual-space non-local information within two parallel pathways that capture pixel-wise contextual information and channel-wise relationships. Unlike \citep{fu2019dual}, which directly append a single attention module on top of FCN for scene segmentation, our NLGM is composed of 5 stacked dual-space non-local blocks (DSNLBs), one at each stage of the top-down pathway. We choose the feature map \(S^5\) extracted from Conv5-3 as the input of the NLGM as it contains high-level semantic information, and yet contains more spatial information than \(S^6\). Fig~\ref{fig:non-local} illustrates the detailed structure of the first DSNLB.

\subsubsection{Spatial-Space Non-Local Block:} For a given feature map \(\mathbf{A}\in \mathds{R}^{C\times H\times W}\), 3 convolutional layers are used to generate 3 distinct feature maps \(\left\{\mathbf{B, C, D}\right\}\in \mathds{R}^{C\times H\times W}\), representing query, key and value respectively. \(\mathbf{B, C}\) and \(\mathbf{D}\) are reshaped to \(\mathds{R}^{C\times K}\), where \(K=H\times W\). A unified similarity matrix \(\mathbf{S}\in \mathds{R}^{K\times K}\) is calculated as:
\begin{equation}
    \mathbf{S}=f(\mathbf{B}^\mathrm{T}\times \mathbf{C}).
\end{equation}
We apply a \(softmax\) normalisation function for $f$ here. The weight matrix \(\mathbf{S}\) models the affinity of features between any two spatial positions. The final weighted feature map \(SN\) in spatial space can be defined as:
\begin{equation}
    SN=(\mathbf{S}^\mathrm{T}\times \mathbf{D})^\mathrm{T}+\mathbf{A},
\end{equation}
where \(SN\in \mathds{R}^{C\times H\times W}\) after reshaping. Informally, \(SN\) may be thought of as an enriched feature representation with global spatial perception, as each value of \(SN\) is a selective weighted sum of all positions across the whole feature map.

\subsubsection{Channel-Space Non-Local Block} Given same feature map \(\mathbf{A}\in \mathds{R}^{C\times H\times W}\), reshaped to \(\mathds{R}^{C\times K}\), the unified similarity matrix \(\mathbf{X}\in \mathds{R}^{C\times C}\) in channel space is calculated as:
\begin{equation}
    \mathbf{X}=f(\mathbf{A}\times \mathbf{A}^\mathrm{T}),
\end{equation}
where \(f\) indicates \(softmax\) function and \(\mathbf{X}\) measures the correlation between any two channels. The weighted feature map \(CN\) in channel space can be defined as:
\begin{equation}
    CN=(\mathbf{X}^\mathrm{T}\times \mathbf{A})+\mathbf{A},
\end{equation}
where \(CN\in \mathds{R}^{C\times H\times W}\) is a feature representation containing long-range dependencies within channel space.

\subsubsection{Non-Local Feature Aggregation} We combine the above features \(SN\) and \(CN\) to generate our non-local feature representations \(N\), exploiting long-range contextual information in both spatial and channel space. \(N\) is given as:
\begin{equation}
    N=D_N(SN+CN),
\end{equation}
where \(D_N\) is a \(1\times 1\) convolutional layer.

\subsubsection{Multi-Hop Communications} Our NLGM is composed of 5 DSNLBs, each generating non-local features \(N^i\) at each stage of top-down pathway. By stacking several DSNLBs, non-local features are progressively refined through multi-hop communication between features that share affinity both channel-wise and spatially. Relevant saliency-specific semantic information is shared across the image space and feature space. Since non-local features have a global view, the challenge of complex scenes and multiple salient objects are better addressed through this improved receptive field.

\subsection{Feature Fusion}\label{subsection2}
\begin{figure}[t]
\begin{center}
\includegraphics[width=1.0\linewidth]{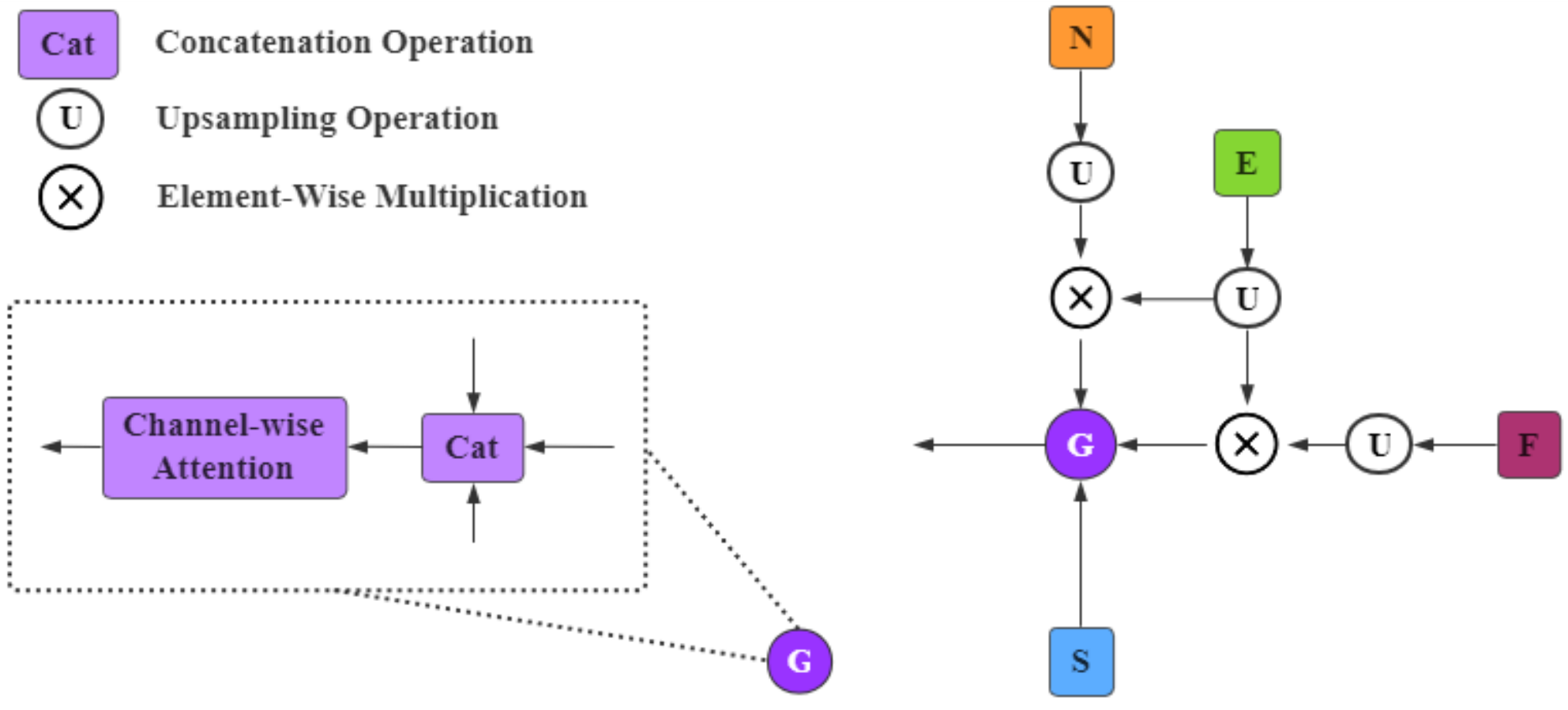}
\end{center}
   \caption{The structure of a feature fusion gate. \(N\), \(E\), \(F\) and \(S\) demonstrate non-local feature, edge feature, saliency feature and the corresponding side output of bottom-up pathway respectively.}
\label{fig:FFG}
\end{figure}
The majority of the existing SOD models fuse different features without distinction. Redundant and mutually incompatible features from different objectives may harm overall performance once combined. As presented in Fig~\ref{fig:FFG}, we leverage channel-wise attention to selectively aggregate different features at each stage of top-down pathway.

\subsubsection{Edge Refinement Module}

Motivated by the commonly used boundary detection in SOD models \citep{qin2019basnet,zhao2019egnet}, salient edge features are modeled, as part of the feature fusion gate, to support the training of non-local and salient features. Edge features are generated from low-level features \(S^2\) and high-level features \(S^5\) through bottom-up process, which are progressively refined within an Edge Refinement Module (ERM). We found that \(S^2\) and \(S^5\) outperformed \(S^1\) and \(S^6\) in this role. \(S^1\) is lower level with a reduced receptive field size, while \(S^6\) is high level, but has limited resolution.

The Edge Refinement Module (ERM) consists of 5 edge refinement blocks (ERBs) aligned with the respective blocks in the NLGM. Each ERB contains a convolutional layer followed by a ReLU function to generate salient edge features \(E^i\) at each stage. As with the salient features, we apply intermediate supervision in the ERM via a cross-entropy loss. A convolutional layer \(D_E^i\) is used to convert the edge feature to single-channel prediction map. The supervision here can be defined as:
\begin{multline}
    \mathcal{L}^i_E\left(E^i, W_{DE}^i\right)=-\sum_{j\in{V^+}}\log Pred\left(y_j=1|E^i;W_{DE}^i\right) \\-\sum_{j\in{V^-}}\log Pred\left(y_j=0|E^i;W_{DE}^i\right), i\in\left[1,5\right],
\end{multline}
where \(V^+\) and \(V^-\) denote the set of salient edge pixels and other pixels respectively. \(W_{DE}\) indicates the parameters of \(D_E^i\). We weight the two components of the above loss by the number of pixels in each class.

\subsubsection{Feature Fusion Gate}

Salient edge features are combined with salient and non-local features separately through element-wise multiplication. This operation serves to emphasise activations that are shared between feature maps, promoting complimentarity between the non-local, saliency and edge features. Features that are aligned between modules will train more quickly, with activations of those that are less relevant to other blocks reduced. The resulting salient and non-local features are then combined using channel-wise attention. Firstly, we unify the spatial size and channel count:
\begin{gather}
    \hat{N^i}=Up(\delta(O(N^i;\theta));S^i), i\in [1,5],\\
    \hat{F^i}=Up(\delta(O(F^i;\theta));S^{i+1}), i\in [2,6],\\
    \hat{E^i}=Up(\delta(O(E^i;\theta));S^i), i\in [1,5],
\end{gather}
where \(Up(*;S^i)\) upsamples \(*\) to be the same size as \(S^i\). \(O(*;\theta)\) denotes a convolutional layer with parameter \(\theta\) and \(\delta\) is a ReLU activation function, which converts the channel number of \(*\) to the channel number of \(S^i\). The fused features \(F^i_{fusion}\), including the incorporation of salient edge features, are calculated as:
\begin{equation}
    F^i_{fusion}=CA(Cat(S^i, \hat{N^i}\otimes \hat{E^i}, \hat{F}^{i+1}\otimes \hat{E^i})),
\end{equation}
where \(\otimes\) denotes the element-wise multiplication. \(Cat\) is the concatenation operation. \(CA\) indicates the channel-wise attention, which can be defined as:
\begin{equation}
    CA(*, \theta_{ca})=*\cdot (\sigma(fc_{2}(\delta(fc_{1}(ap(*,\theta_{1}))),\theta_{2}))),
\end{equation}
where \(\theta_{ca}\) denotes the parameters in channel-wise attention, \(ap\) is a global average pooling layer and \(fc\) is a fully-connected layer. \(\sigma\) and \(\delta\) refer to the sigmoid function and ReLU functions respectively. The feature fusion gate here provides a mechanism to select the most useful channels for saliency from each module, fusing features with distinction.

\begin{figure}[t]
\begin{center}
\includegraphics[width=1.0\linewidth]{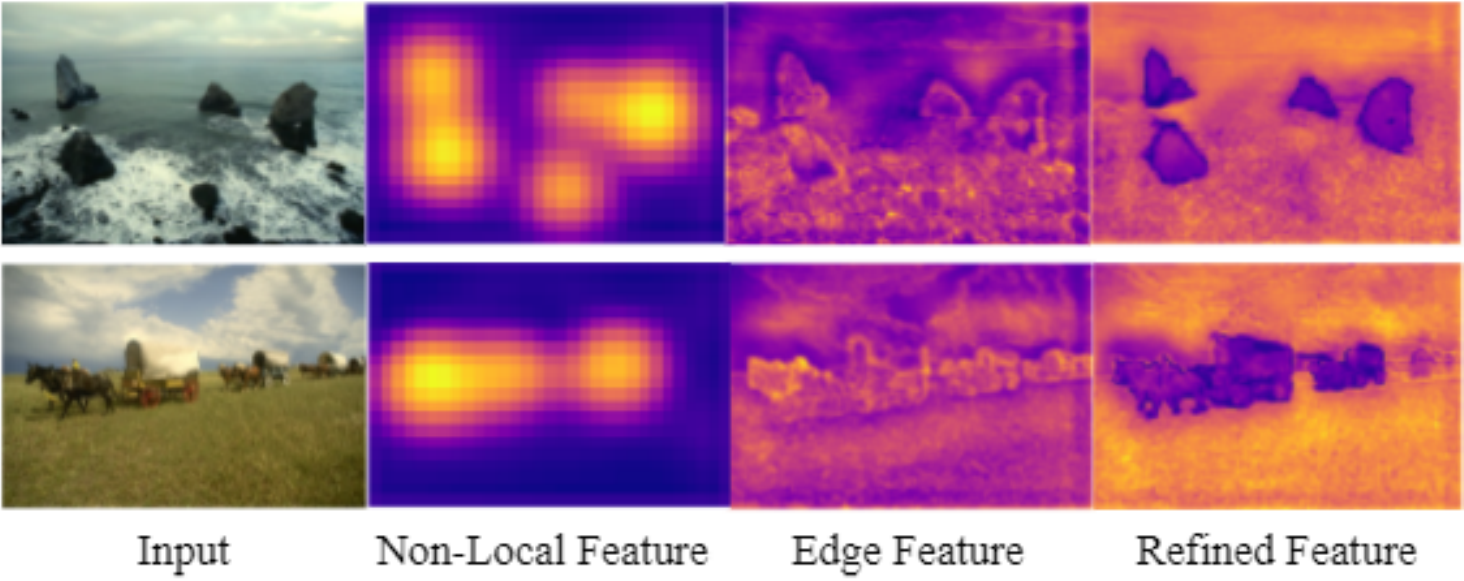}
\end{center}
   \caption{Feature visualization of non-local features, edge features and the refined features after feature fusion. As can be seen, our non-local features highlight objects across the scene. After the feature fusion, multiple salient objects are clearly defined.}
\label{fig:featureVisualizartion}
\end{figure}

\subsection{Saliency Inference}\label{subsection3}
To make full use of the multi-scale saliency features, we hierarchically generate the final prediction map based on the fusion of six saliency features \(F^i\) from coarse-to-fine manner. This multi-scale fusion strategy also serves to reduce the risk of missing salient objects across multi-saliency visual scenes. Complementary features \(F^2, F^3, F^4, F^5, F^6\) are upsampled and convolved to have equal spatial and feature size to \(F^1\), they are then combined using element-wise addition to produce a final feature \(F_{fin}\). A convolutional layer \(D_{P}\) is used to convert the feature map \(F_{fin}\) to a single-channel prediction map, trained using cross entropy: 
\begin{multline}
    \mathcal{L}_{fin}\left(F_{fin}, W_{P}\right)=-\sum_{j\in{Y^+}}\log Pred\left(y_j=1|F_{fin};W_{P}\right) \\-\sum_{j\in{Y^-}}\log Pred\left(y_j=0|F_{fin};W_{P}\right), 
\end{multline}
where \(Y^{+}\) and \(Y^-\) denote the set of salient pixels and non-salient pixels respectively. \(W_{P}\) refers to the parameters of convolutional layer \(D_{P}\). Therefore the total loss \(\mathds{L}\) in our proposed method can be denoted as:
\begin{multline}
    \mathds{L}=\sum_{i=1}^{6}\mathcal{L}_F^i(F^i;W_{DF}^i)+\sum_{i=1}^{5}\mathcal{L}_E^i(E^i;W_{DE}^i)\\
    +\mathcal{L}_{fin}\left(F_{fin}, W_{P}\right)
\end{multline}

%-------------------------------------------------------------------------
\section{Experiments}
\subsection{Datasets:}
\begin{figure}[h]
\begin{center}
\includegraphics[width=1.0\linewidth]{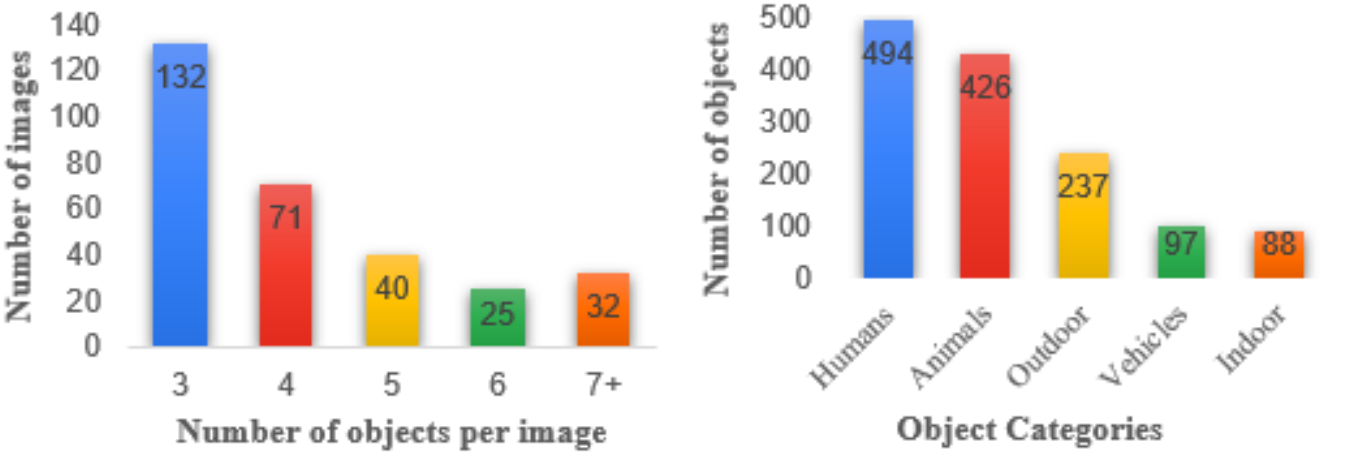}
\end{center}
   \caption{The distribution of the proposed MSOD dataset.}
\label{fig:dataset}
\end{figure}

To demonstrate the performance of our proposed method, we evaluate our model on five widely used benchmark datasets: DUT-OMRON \citep{yang2013saliency}, HKU-IS \citep{li2015visual}, DUTS \citep{wang2017learning}, ECSSD \citep{yan2013hierarchical} and SOD \citep{movahedi2010design}. Scenes including multiple salient objects do exist in each dataset, but the frequency varies from a minimum of 9.8\% for ECSSD to 50.3\% for HKU-IS, with an average across the datasets of 28\%. Of these multi-object images, the majority contain only two objects. Metrics calculated across these datasets will respond to model performance specific to multiple objects, but evaluation is challenging. Here, we curate a new dataset, MSOD, containing the most challenging multi-object scenes across the five datasets. In order to evaluate our, and other methods on complex multi-object images, scenes were only included if they contained three or more salient objects. The total MSOD dataset comprises 300 test images with 1342 total objects. The number of objects in each image varies from 3 to 19 and the distribution is shown in Fig~\ref{fig:dataset}. The dataset comprises a variety of object classes, and a varied number of these objects across the image. We observed that the objects did not exhibit any particular predictable spatial locations.

\subsection{Evaluation Metrics:}
Three widely used metrics are applied for performance evaluation: F-measure, mean absolute error (MAE) and the structure-based metric S-measure \citep{fan2017structure}. F-measure is a weighted harmonic mean of precision and recall, defined as:
\begin{equation}
    F_{\beta}=\frac{(1+\beta^2)\times Precision\times Recall}{\beta^2 \times Precision+Recall}
\end{equation}
where \(\beta^2\) is commonly set to 0.3 in previous works to emphasize precision. Following most state-of-the-art SOD approaches, we report the maximum \(F_\beta\) from all pairs of precision and recall on different thresholds. We also use MAE to compare the prediction map \(P\) and ground truth \(Y\), of size \(W\times H\), defined as:
\begin{equation}
    MAE=\frac{1}{W\times H}\sum_{i=1}^{W}\sum_{j=1}^{H}{|P(i,j)-Y(i,j)|}
\end{equation}
Finally, we use S-measure to consider both the region-aware \(S_r\) and object-aware \(S_o\) structural similarity, defined as:
\begin{equation}
    S=\alpha\times S_0+\left(1-\alpha\right)\times S_r
\end{equation}
where \(\alpha\) is set to 0.5 by default. More details about S-measure can be found in \citep{fan2017structure}.

\begin{table*}[t]
\begin{center}
\scalebox{0.61}{
\begin{tabular}{l|ccc|ccc|ccc|ccc|ccc|ccc}
\toprule[1.5pt]
\multicolumn{1}{c}{}                        & \multicolumn{3}{|c|}{\textbf{ECSSD} }                                                                                            & \multicolumn{3}{c|}{\textbf{DUTS-TE}}                                                                                          & \multicolumn{3}{c|}{\textbf{HKU-IS}}                                                                                           & \multicolumn{3}{c|}{\textbf{DUT-O}}                                                                                            & \multicolumn{3}{c|}{\textbf{SOD}}     & \multicolumn{3}{c}{\textbf{MSOD}}   

\\\multicolumn{1}{c}{}                        & \multicolumn{3}{|c|}{1000 images}                                                                                            & \multicolumn{3}{c|}{5019 images}                                                                                          & \multicolumn{3}{c|}{1447 images}                                                                                           & \multicolumn{3}{c|}{5168 images}                                                                                            & \multicolumn{3}{c|}{300 images}     & \multicolumn{3}{c}{300 images}

\\ \cmidrule(lr){2-4}  \cmidrule(lr){5-7}\cmidrule(lr){8-10}\cmidrule(lr){11-13}\cmidrule(lr){14-16}\cmidrule(lr){17-19}
\multicolumn{1}{c|}{\multirow{-2}{*}{Model}} & MaxF \(\uparrow\)                     & MAE \(\downarrow\)                    & S \(\uparrow\)                        & MaxF \(\uparrow\)                     & MAE \(\downarrow\)                    & S \(\uparrow\)                        & MaxF \(\uparrow\)                     & MAE \(\downarrow\)                    & S \(\uparrow\)                        & MaxF \(\uparrow\)                     & MAE \(\downarrow\)                    & S \(\uparrow\)                        & MaxF \(\uparrow\)                     & MAE \(\downarrow\)                    & S \(\uparrow\)          & MaxF \(\uparrow\)                     & MAE \(\downarrow\)                    & S \(\uparrow\)               \\ \midrule[1.5pt]
\multicolumn{19}{c}{\textbf{VGG-Backbone}}                                                                                                                                                                                                                                                                                                                                                                                                                                                                                                                                                                                                                          \\ \midrule[1.5pt]
\textbf{DSS (CVPR2017)}                                  & .9207                                 & .0517                                 & .8821                                 & .8251                                 & .0565                                 & .8237                                 & .9161                                 & .0401                                 & .8783                                 & .7812                                 & .0628                                 & .7899                                 & .8410                                 & .1201                                 & .7478                & .8240   & .0550   & .7806                   \\ \midrule
\textbf{BDMP (CVPR2018)}                                 & .9284                                 & .0446                                 & .9109                                 & .8514                                 & .0490                                 & .8616                                 & .9205                                 & .0389                                 & .9065                                 & .7739                                 & .0636                                 & .8091                                 & .8517                                 & .1057                                 & .7833                  & .8401   & .0538   & .8379                   \\ \midrule
\textbf{PAGR (CVPR2018)}                                 & .9259                                 & .0608                                 & .8883                                 & .8540                                 & .0555                                 & .8383                                 & .9187                                 & .0475                                 & .8891                                 & .7706                                 & .0709                                 & .7751                                 & .8358                                 & .1447                                 & .7137        & .8204   & .0627   & .7852                             \\ \midrule
%\textbf{PiCANet (CVPR2018)}                            & .9313                                 & .0464                                 & .9138                                 & .8512                                 & .0541                                 & .8607                                 & .9216                                 & .0415                                 & .9054                                 & .7938                                 & .0679                                 & .8263                                 & .8504                                 & .1010                                 & .7867                   & .8168   & .0618   & .8246                  \\ \midrule
\textbf{RAS (ECCV2018)}                                  & .9211                                 & .0564                                 & .8928                                 & .8311                                 & .0594                                 & .8385                                 & .9128                                 & .0454                                 & .8874                                 & .7864                                 & .0617                                 & .8141                                 & .8473                                 & .1225                                 & .7608                & .8370   & .0597   & .8167                     \\ \midrule
\textbf{BASNet (CVPR2019)}                               & .9425                                 & .0370                                 & .9162                                 & .8594                                 & .0476                                 & .8656                                 & .9297                                 & .0329                                 & .9077                                 & .8052                                 & .0565                                 & .8361                                 & .8487                                 & .1119                                 & .7660                  & .8396   & .0541   & .8306                   \\ \midrule
\textbf{AFNet (CVPR2019)}                                & .9350                                 & .0418                                 & .9134                                 & .8628                                 & .0458                                 & .8666                                 & .9252                                 & .0355                                 & .9058                                 & .7970                                 & .0573                                 & .8258                                 & .8499                                 & .1087                                 & .7700               & .8276   & .0547   & .8191                      \\ \midrule

\textbf{Ours}                                 & .9485 & \textcolor{cyan}{ \textbf{.0344}} & .9261 & \textcolor{blue}{ \textbf{.8894}} & .0381 & \textcolor{blue}{ \textbf{.8878}} &.9350 & \textcolor{cyan}{ \textbf{.0300}} & .9167 & \textcolor{blue}{ \textbf{.8194}} & .0541 & \textcolor{blue}{ \textbf{.8421}} & \textcolor{cyan}{ \textbf{.8761}} & .0996 &.7922   &.8531   &.0478  & .8410     \\ \midrule[1.5pt]
\multicolumn{19}{c}{\textbf{ResNet-Backbone}}                                                                                                                                                                                                                                                                                                                                                                                                                                                                                                                                                                                                                       \\ \midrule[1.5pt]
\textbf{PiCANet (CVPR2018)}                            & .9349                                 & .0464                                 & .9170                                 & .8597                                 & .0506                                 & .8686                                 & .9193                                 & .0437                                 & .9045                                 & .8027                                 & .0653                                 & .8318                                 & .8528                                 & .1024                                 & .7871                           & .8190   & .0641   & .8223         \\ \midrule
\textbf{PoolNet (CVPR2019)}                            & \textcolor{cyan}{\textbf{.9489}}                                 & .0350                                 & \textcolor{cyan}{\textbf{.9263}}                                 & \textcolor{cyan}{ \textbf{.8891}}                                 & \textcolor{blue}{ \textbf{.0368}}                                 & .8865                                 & \textcolor{blue}{ \textbf{.9358}}                                 & \textcolor{cyan}{ \textbf{.0300}}                                 & \textcolor{cyan}{ \textbf{.9187  }}                               & .8048                                 & \textcolor{cyan}{ \textbf{.0539}}                                 & .8309                                 & .8706                                 & .1034                                 & .7854                             & \textcolor{cyan}{ \textbf{.8546}}   & \textcolor{blue}{ \textbf{.0459}}   & \textcolor{cyan}{ \textbf{.8429}}        \\ \midrule
\textbf{EGNet (ICCV2019)}                              & .9474                                 & .0374                                 & .9247                                 & .8885                                 & .0392                                 & \textcolor{cyan}{ \textbf{.8868}}                                 & .9352                                 & .0309                                 & .9179                                 & .8152                                 & \textcolor{blue}{ \textbf{.0531}}                               &\textcolor{cyan}{ \textbf{.8408}}                                 &\textcolor{blue}{ \textbf{.8778}}                                & \textcolor{cyan}{ \textbf{.0969}}                                 & \textcolor{blue}{ \textbf{.8000}}                              & .8516   & \textcolor{cyan}{ \textbf{.0470}}   & .8402       \\ \midrule
\textbf{CPD (CVPR2019)}                                & .9393                                 & .0371                                 & .9181                                 & .8653                                 & .0434                                 & .8689                                 & .9252                                 & .0339                                 & .9064                                 & .7964                                 & .0560                                 & .8247                                 & .8568                                 & .1095                                 & .7646                              & .8241   & .0539   & .8109       \\ \midrule
\textbf{SCRN (ICCV2019)}                                 &\textcolor{blue}{ \textbf{.9496}}                                 & .0375                                 & \textcolor{blue}{ \textbf{.9272}}                                 & .8875                                 & .0398                                 & .8847                                 & .9351                                 & .0332                                 & .9169                                 & .8112                                 & .0560                                 & .8364                                 & .8655                                 & .1046                                 & .7851                             &.8384   & .0527   & .8244        \\ \midrule
\textbf{GateNet (ECCV2020)}                                & .9454                                 & .0401                                 & .9198                                 & .8873                                 & .0401                                 & .8847                                 & .9334                                 & .0331                                 & .9153                                 &\textcolor{cyan}{ \textbf{.8178}}                                & .0549                                 & .8380                                 & .8731                                 & .0981                                 & .7948                          & \textcolor{blue}{ \textbf{.8623}}  &.0483   &\textcolor{blue}{ \textbf{.8507}}          \\ \midrule
\textbf{MINet (CVPR2020)}                                & .9475                                 &\textcolor{blue}{ \textbf{.0335}}                                  & .9249                                 & .8836                                 &\textcolor{cyan}{ \textbf{ .0372 }}                                & .8837                                 &\textcolor{cyan}{ \textbf{ .9353 }}                                & \textcolor{red}{ \textbf{.0283}}                                 &\textcolor{blue}{ \textbf{.9197}}                                 & .8097                                 & .0555                                 & .8325                                 & .8730                                 & \textcolor{red}{ \textbf{.0905}}                                 & \textcolor{cyan}{ \textbf{.7973}}                               & .8472   & .0474   & .8400      \\ \midrule

\textbf{SCWSSOD* (AAAI2021)}                                & .9145                                 &.0489                                  & .8818                                 & .8440                                 & .0487                                 & .8405                                 & .9111                                 & .0375                                 &.8824                                 & .7823                                 & .0602                                 & .8117                                 & .8367                                 & .1077                                & .7503                               & .8329   & .0534   & .8060      \\ \midrule

\textbf{Ours}                                 & \textcolor{red}{ \textbf{.9519}} & \textcolor{red}{ \textbf{.0325}} & \textcolor{red}{ \textbf{.9297}} & \textcolor{red}{ \textbf{.8967}} & \textcolor{red}{ \textbf{.0358}} & \textcolor{red}{ \textbf{.8946}} & \textcolor{red}{ \textbf{.9389}} & \textcolor{blue}{ \textbf{.0293}} & \textcolor{red}{ \textbf{.9218}} & \textcolor{red}{ \textbf{.8234}} & \textcolor{red}{ \textbf{.0530}} & \textcolor{red}{ \textbf{.8470}} & \textcolor{red}{ \textbf{.8786}} & \textcolor{blue}{ \textbf{.0934}} & \textcolor{red}{ \textbf{.8024}} &\textcolor{red}{ \textbf{.8720}}   & \textcolor{red}{ \textbf{.0442}} &  \textcolor{red}{ \textbf{.8614} }\\ \bottomrule[1.5pt]
\end{tabular}}
\end{center}
\caption{Quantitative comparison with other state-of-the-art methods on 5 widely used datasets and the proposed MSOD dataset. \(\uparrow\) and \(\downarrow\) indicate higher or lower is better respectively and * denotes weakly-supervised methods. The best three results among both backbones are marked as \textcolor{red}{ \textbf{red}}, \textcolor{blue}{ \textbf{blue}} and \textcolor{cyan}{ \textbf{cyan}}.  Our method achieves top results under 3 evaluation metrics across all datasets without any pre-processing and post-processing.}
\label{tab:quantativeComparison}
\end{table*}

\begin{table}[t]
\begin{center}
\scalebox{0.68}{
\begin{tabular}{ccc|ccc|ccc}
\toprule[1.5pt]
\multicolumn{3}{c|}{Models} & \multicolumn{3}{c}{DUTS-TE } & \multicolumn{3}{c}{DUT-OMRON } \\ \cmidrule(lr){1-3} \cmidrule(lr){4-6} \cmidrule(lr){7-9}
NLGM     & FFG     & ERM    & MaxF \(\uparrow\)    & MAE \(\downarrow\)     & S \(\uparrow\)   & MaxF \(\uparrow\)     & MAE \(\downarrow\)      & S    \(\uparrow\)    \\ \midrule[1.5pt]
         &         &         & .8761    & .0425   & .8750   & .7958    & .0575    & .8276    \\ \midrule
\checkmark         &         &         & .8836    & .0410   & .8809   & .8103    & .0562    & .8357    \\ \midrule
\checkmark         &      &\checkmark         & .8847    & .0410   & .8839   & .8138    & .0563    & .8366    \\ \midrule
\checkmark         &   \checkmark      &         & .8858    & .0396   & .8857   & .8153    & .0560    & .8379    \\ \midrule
\checkmark        &  \checkmark       &   \checkmark      &\textcolor{red}{\textbf{.8894}}     &\textcolor{red}{\textbf{.0381}}    &\textcolor{red}{\textbf{.8878}}    &\textcolor{red}{\textbf{.8194}}     &\textcolor{red}{\textbf{.0541}}     &\textcolor{red}{\textbf{.8421}}     \\ 
\bottomrule[1.5pt]
\end{tabular}
}
\end{center}
\caption{Ablation analysis of different components in our proposed architecture.}
\label{tab:Ablationanalysis}
\end{table}

\begin{table}[t]
\begin{center}

\scalebox{0.69}{
\begin{tabular}{l|ccc|ccc}
\toprule[1.5pt]
\multicolumn{1}{c|}{\multirow{2}{*}{NLGM Configurations}} & \multicolumn{3}{c}{DUTS-TE } & \multicolumn{3}{c}{DUT-OMRON } \\ \cmidrule(lr){2-4} \cmidrule(lr){5-7}
\multicolumn{1}{c|}{}                        & MaxF \(\uparrow\)    & MAE \(\downarrow\)    & S \(\uparrow\)      & MaxF    \(\uparrow\) & MAE \(\downarrow\)     & S \(\uparrow\)       \\ \midrule[1.5pt]
SSNLB                          & .8813    & .0411   & .8789   & .8081    & .0563    & .8336    \\ \midrule
CSNLB                          & .8816    & .0415   & .8790   & .8079    & .0566    & .8348    \\ \midrule
SSNLB+CSNLB                    &\textcolor{red}{\textbf{.8836}}     &\textcolor{red}{\textbf{.0410}}    &\textcolor{red}{\textbf{.8809}}    &\textcolor{red}{\textbf{.8103}}     &\textcolor{red}{\textbf{.0562}}      &\textcolor{red}{\textbf{.8357}}     \\ \bottomrule[1.5pt]
\end{tabular}}
\end{center}
\caption{Performance comparison of different NLGM configurations. SSNLB and CSNLB refer to spatial-space non-local block and channel-space non-local block respectively. All three configurations are without FFG and ERM.}
\label{tab:NLGMconfigurations}
\end{table}

\subsection{Implementation Details:} The proposed approach is implemented in PyTorch and trained on the DUTS-TR \citep{wang2017learning} dataset, where the salient edge ground truth is calculated by sober operator. To compare our method against other state-of-the-art methods, we train our model using both VGG \citep{simonyan2014very} and Resnet-50 \citep{he2016deep} backbones. The parameters of backbones are initialized using pretrained models on ImageNet \citep{krizhevsky2012imagenet}, while the weights of newly added layers are initialized randomly. We use the Adam \citep{kingma2014adam} optimizer with an initial learning rate of 2e-5, which is divided by 10 after 30 epochs. Our model is trained for 40 epochs in total, which tipically takes 3 days in a single 2080Ti with a forward pass taking 0.02s.
\subsection{Comparisons with the State-of-the-Art}
We compare our proposed method against 14 recent state-of-the-art methods: DSS \citep{hou2017deeply}, BDMP \citep{zhang2018bi}, PAGR \citep{zhang2018progressive}, PiCANet \citep{liu2018picanet}, RAS \citep{chen2018reverse}, BASNet \citep{qin2019basnet}, AFNet \citep{feng2019attentive}, PoolNet \citep{liu2019simple}, EGNet \citep{zhao2019egnet}, CPD \citep{wu2019cascaded}, SCRN \citep{wu2019stacked}, GateNet \citep{zhao2020suppress}, MINet \citep{pang2020multi} and SCWSSOD \citep{yu2021structure}. For fair comparison, all the saliency maps of competing methods were produced by either pre-trained models, or pre-generated by the authors.

\subsubsection{Quantitative Comparison:} Table~\ref{tab:quantativeComparison} shows that our proposed method achieves the best result under 3 evaluation metrics across the datasets. Against the current top approaches: MINet \citep{pang2020multi}, PoolNet \citep{liu2019simple}, EGNet \citep{zhao2019egnet} and SCRN \citep{wu2019stacked}, the average improvement of our proposed method on 5 widely used datasets is 0.92\%, 2.27\%, 2.43\% and 4.09\% respectively. A further performance improvement can be observed in MSOD dataset, with the average improvement over these approaches increasing to 4.08\%, 2.64\%, 3.63\% and 8.21\% respectively. This demonstrates the strong multiple salient object detection ability of our proposed method, which obtains state-of-the-art performance on this challenging dataset.

\subsubsection{Precision-Recall Curves:} 
\begin{figure*}[t]
\centering
\subfigure[DUT-OMRON]{
\begin{minipage}[t]{0.24\linewidth}
\centering
\includegraphics[width=1.70in]{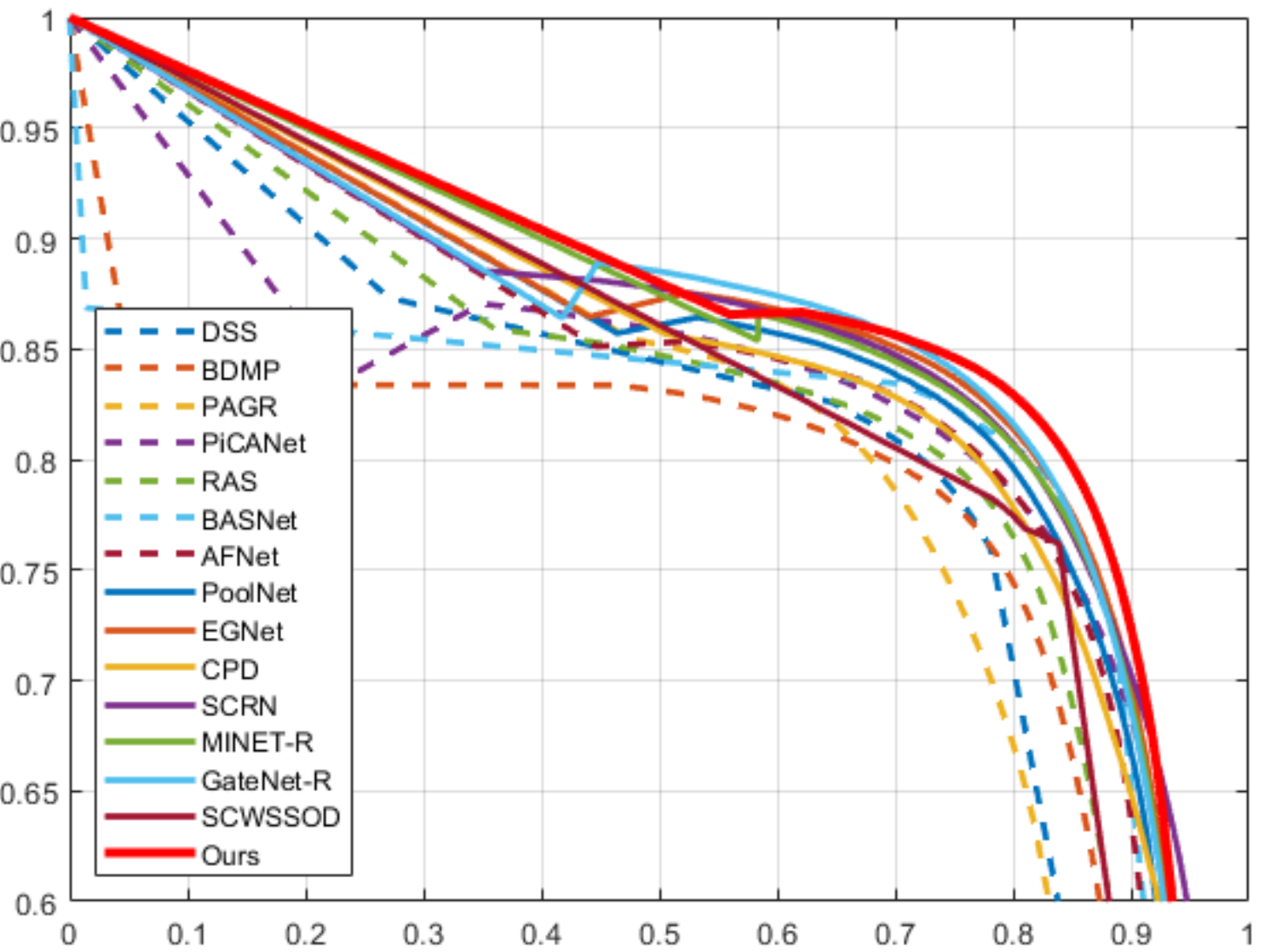}
%\caption{fig1}
\end{minipage}%
}%
\subfigure[DUTS-TE]{
\begin{minipage}[t]{0.24\linewidth}
\centering
\includegraphics[width=1.70in]{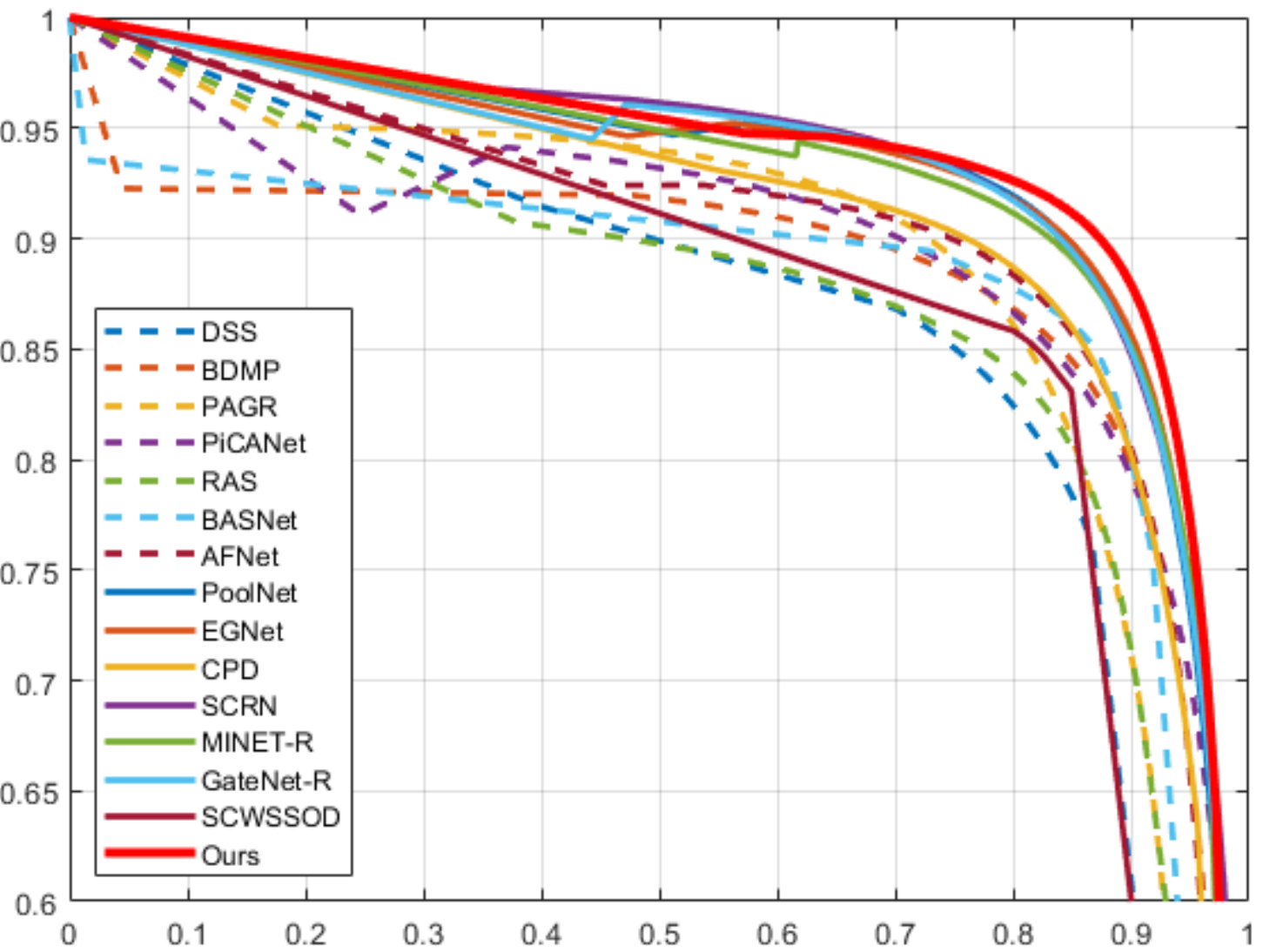}
%\caption{fig2}
\end{minipage}%
}%
\subfigure[ECSSD]{
\begin{minipage}[t]{0.24\linewidth}
\centering
\includegraphics[width=1.70in]{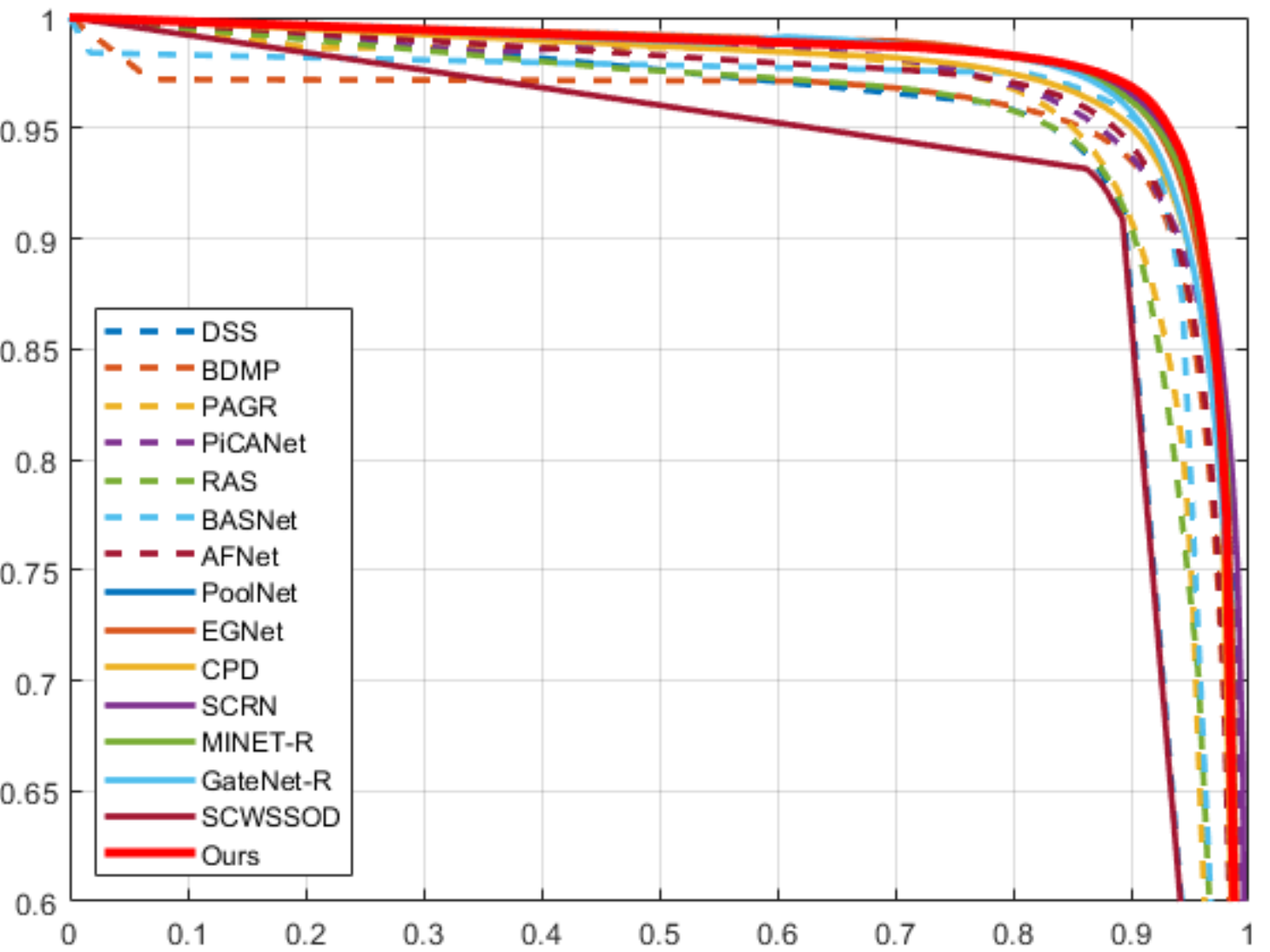}
%\caption{fig2}
\end{minipage}
}%
\subfigure[MSOD]{
\begin{minipage}[t]{0.24\linewidth}
\centering
\includegraphics[width=1.70in]{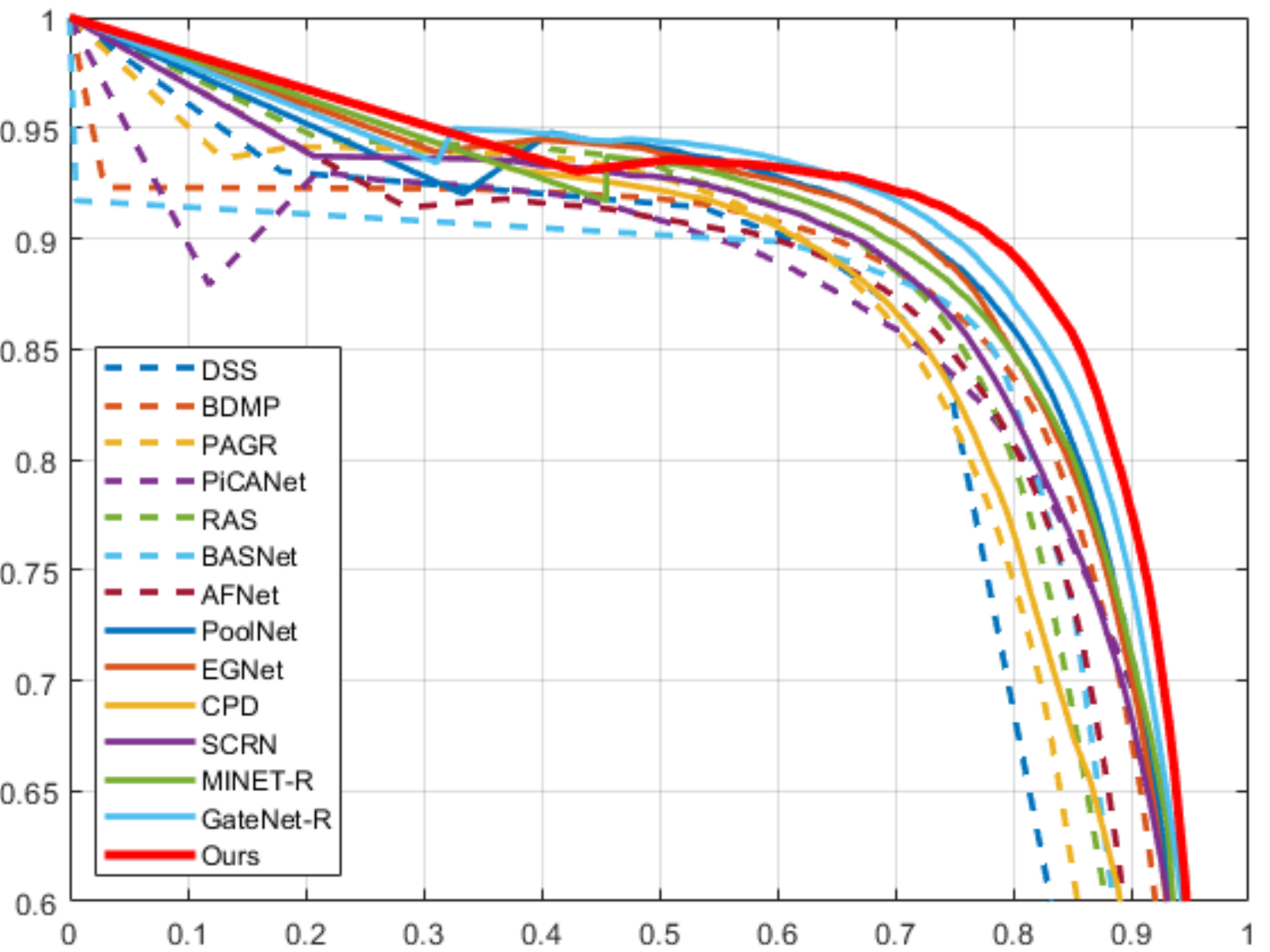}
%\caption{fig2}
\end{minipage}
}%
\centering
\caption{Precision (vertical axis) recall (horizontal axis) curves on three popular salient object detection datasets and the proposed MSOD dataset. The red solid line demonstrates our proposed method.}
\label{fig:PRCurve}
\end{figure*}

We plot precision-recall curves over 3 popular SOD datasets and the proposed MSOD dataset in Fig~\ref{fig:PRCurve}. Our proposed method outperforms all other approaches at most thresholds, especially in the two largest datasets DUT-OMRON and DUTS-TE.

\subsubsection{Visual Comparison:} 
\begin{figure*}[t]
\begin{center}
\includegraphics[width=1\linewidth]{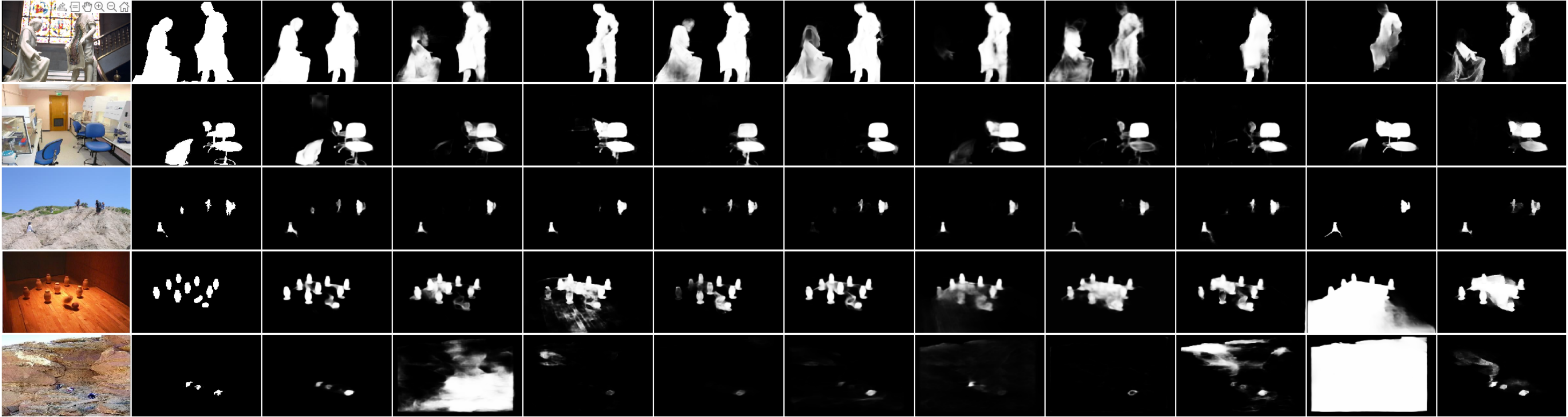}
\centering
\begin{minipage}[t]{0.083\linewidth}
\centering
\footnotesize Image
\end{minipage}%
\begin{minipage}[t]{0.083\linewidth}
\centering
\footnotesize GT
\end{minipage}%
\begin{minipage}[t]{0.083\linewidth}
\centering
\footnotesize Ours
\end{minipage}%
\begin{minipage}[t]{0.083\linewidth}
\centering
\footnotesize GateNet
\end{minipage}%
\begin{minipage}[t]{0.083\linewidth}
\centering
\footnotesize MINet
\end{minipage}%
\begin{minipage}[t]{0.083\linewidth}
\centering
\footnotesize PoolNet
\end{minipage}%
\begin{minipage}[t]{0.083\linewidth}
\centering
\footnotesize EGNet
\end{minipage}%
\begin{minipage}[t]{0.083\linewidth}
\centering
\footnotesize SCRN
\end{minipage}%
\begin{minipage}[t]{0.083\linewidth}
\centering
\footnotesize PAGR
\end{minipage}%
\begin{minipage}[t]{0.083\linewidth}
\centering
\footnotesize RAS
\end{minipage}%
\begin{minipage}[t]{0.083\linewidth}
\centering
\footnotesize BASNet
\end{minipage}%
\begin{minipage}[t]{0.083\linewidth}
\centering
\footnotesize AFNet
\end{minipage}%

\end{center}
\caption{Qualitative comparisons with state-of-the-art approaches over some of the challenging images.}
\label{fig:qualitativeComparisons}
\end{figure*}

Qualitative results of our method can be seen in Fig~\ref{fig:qualitativeComparisons}, where our method provides excellent performance in multi-saliency images. Our use of non-local features and top-down feature fusion can make full use of the long-range dependencies between salient objects, with information shared between separate regions of the image.

\subsection{Ablation Studies}
In this section we investigate the contribution of different components in our proposed approach. All experiments are based on VGG backbone and two largest datasets: DUTS-TE and DUT-OMRON.

\subsubsection{Effectiveness of NLGM:} Compared to a baseline structure (\nth{1} row in Table~\ref{tab:Ablationanalysis}), non-local guidance (\nth{2} row) provides performance gains across all evaluation metrics on both datasets. This demonstrates the effectiveness of capturing long-range dependencies in NLGM.

\subsubsection{Effectiveness of FFG:} Incorporating feature fusion (the \nth{4} and \nth{5} row in Table~\ref{tab:Ablationanalysis}) can also improve the performance beyond the ones without FFG across all metrics on both datasets, indicating the importance of filtering the redundant information from each part of the architecture.

\subsubsection{Effectiveness of NLGM~\&~ERM:} Compared to \nth{2} row, the \nth{3} row in Table~\ref{tab:Ablationanalysis} has a higher performance, which proves the effectiveness of introducing both NLGM and ERM. Features that are mutually beneficial to both non-local and edge saliency are emphasised. These modules appear to complement each other, with non-local, edge and saliency all contributing to the accurate recovery of salient regions.

\subsubsection{Effectiveness of FFG~\&~ERM:} By introducing edge features as part of the FFG, performance gains can be observed (\nth{5} row and \nth{4} row in Table~\ref{tab:Ablationanalysis}), which demonstrates that the edge refinement module effectively promotes relevant salient features prior to their combination within the FFG.

% \begin{figure*}
% \begin{center}

% \subfigure[DUT-OMRON\citep{yang2013saliency}]{
% \begin{minipage}[t]{0.20\linewidth}
% \centering
% \includegraphics[width=2cm]{DUTOMRON.pdf}
% \end{minipage}%
% }%
% \subfigure[DUTS-TE\citep{wang2017learning}]{
% \begin{minipage}[t]{0.20\linewidth}
% \centering
% \includegraphics[width=2cm]{DUTS.pdf}
% \end{minipage}%
% }%
% \subfigure[ECSSD\citep{yan2013hierarchical}]{
% \begin{minipage}[t]{0.20\linewidth}
% \centering
% \includegraphics[width=2cm]{ECSSD.pdf}
% \end{minipage}
% }%

% \subfigure[MSOD]{
% \begin{minipage}[t]{0.20\linewidth}
% \centering
% \includegraphics[width=2cm]{MSOD.pdf}
% \end{minipage}
% }%

% \end{center}
%   \caption{Precision (vertical axis) recall (horizontal axis) curves on three popular salient object detection datasets. The red solid line demonstrates our proposed method.}
% \label{fig:PRCurve}
% \end{figure*}

\subsubsection{Architectures of NLGM:} 
\begin{figure}[t]
\begin{center}
\includegraphics[width=0.88\linewidth]{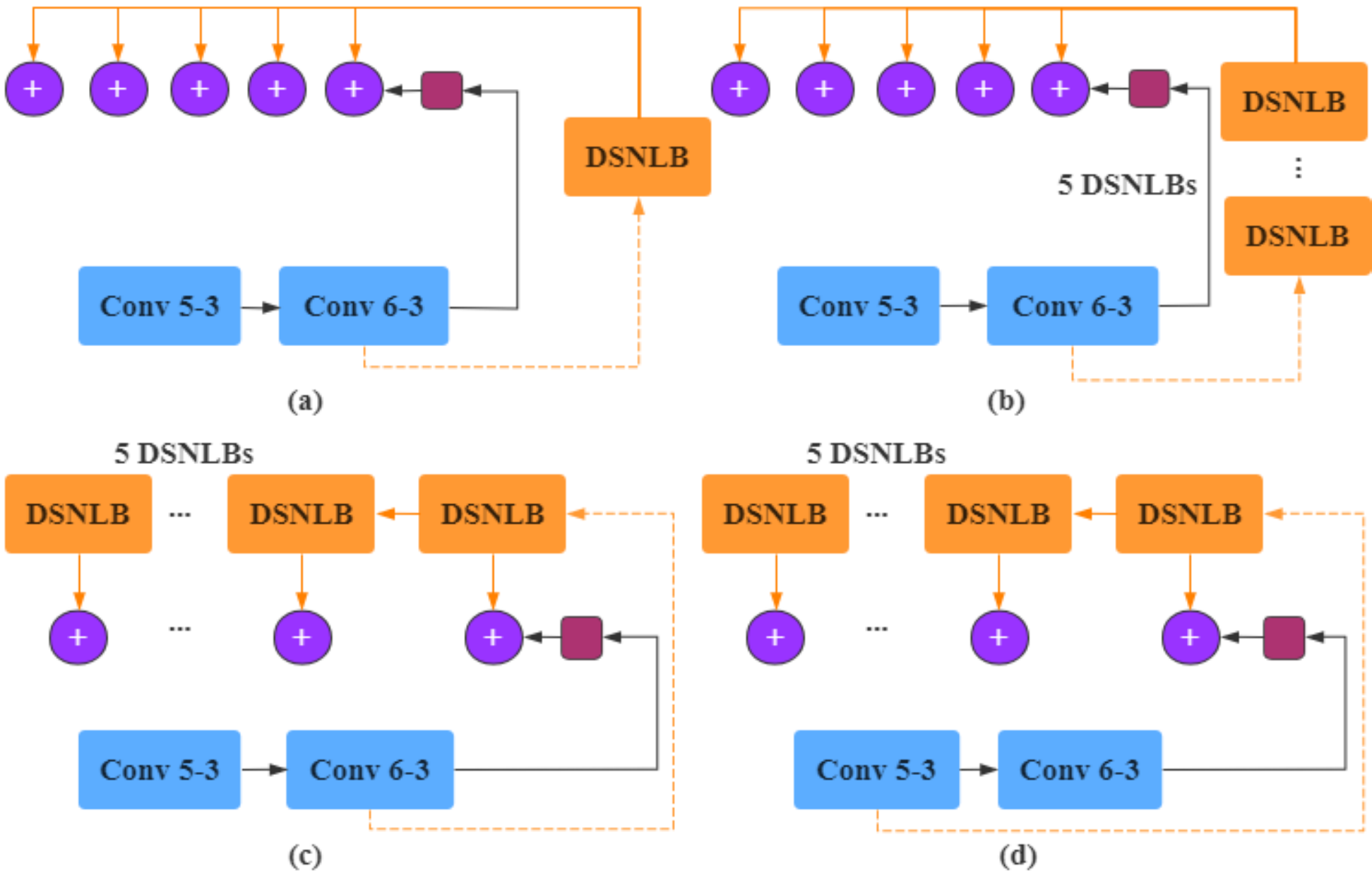}
\end{center}
   \caption{Different architectures of NLGM. All structures here are without FFG and ERGM. Element-wise addition operation is used at each stage to fuse different features.}
\label{fig:NLBSettings}
\end{figure}

We perform experiments to explore the effect of different structures of NLGM. We evaluate four different architectures to incorporate non-local information into our U-shape network. The architectures and the corresponding performance are shown in Fig~\ref{fig:NLBSettings} and Table~\ref{tab:NLGMarchitectures}. Our baseline model (a) utilises a single DSNLB drawing features from \(S^6\), with the same output distributed to all top-down stages. Model (b) incorporates a stack of 5 DSNLBs in this same construction, evaluating the effect of chains of non-local blocks over single instances. We find that a stack of DSNLBs improve performance on all metrics, indicating the effectiveness of long-range multi-hop communications that build richer relevant salient features across both spatial and channel space. Model (c) draws features again from \(S^6\) but distributes the DSNLBs throughout the top-down pathway as per our main architecture in fig~\ref{fig:overallpipepline}. Performance is again improved a little, likely because this one-to-one guidance method can generate adaptive non-local features appropriate to each scale of saliency features \(F\). Finally, model (d) draws features from \(S^5\) as our final architecture does. The improvement suggests that the increased spatial size of these features is better exploited by the DSNLBs over the relatively small spatial size of that in \(S^6\).
%We perform experiments to explore the effect of different structures of NLGM. We evaluated four different architectures to incorporate non-local information into our U-shape network. The architectures and the corresponding performance are shown in fig~\ref{fig:NLBSettings} and Table~\ref{tab:NLGMarchitectures} respectively. Our baseline model (a) utilise a single DSNLB drawing features from \(S^6\). Model (b) incorporates a stack of 5 DSNLBs in this same construction. We find that a stack of DSNLBs improve performance on all metrics, indicating the effectiveness of long-range multi-hop communications. Model (c) instead distributes the DSNLBs throughout the top-down pathway. Performance is improved a little, likely because this one-to-one guidance method can generate adaptive non-local features relevant to each scale. Finally, model (d) draws features from \(S^5\) as our final architecture does. The improvement suggests that the increased spatial size of these features is preferable.

\begin{table}[t]
\begin{center}
\scalebox{0.69}{\begin{tabular}{c|ccc|ccc}
\toprule[1.5pt]
\multicolumn{1}{c|}{\multirow{2}{*}{NLGM Architectures}} & \multicolumn{3}{c}{DUTS-TE } & \multicolumn{3}{c}{DUT-OMRON } \\ \cmidrule(lr){2-4} \cmidrule(lr){5-7}
\multicolumn{1}{c|}{}                        & MaxF \(\uparrow\)    & MAE \(\downarrow\)    & S \(\uparrow\)      & MaxF    \(\uparrow\) & MAE \(\downarrow\)     & S \(\uparrow\)       \\ \midrule[1.5pt]
(a)                      & .8796    & .0419   & .8787   & .8052    & .0567    & .8313    \\ \midrule
(b)                    & .8815    & .0415   & .8801   & .8075    & .0565    & .8330    \\ \midrule
(c)                & .8816    & .0417   & .8805   & .8083    & .0565    & .8345    \\ \midrule
(d)               & \textcolor{red}{\textbf{.8836}}    & \textcolor{red}{\textbf{.0410}}   & \textcolor{red}{\textbf{.8809}}   & \textcolor{red}{\textbf{.8103}}    &\textcolor{red}{\textbf{.0562}}     & \textcolor{red}{\textbf{.8357}}    \\ \bottomrule[1.5pt]
\end{tabular}}
\end{center}
\caption{Performance comparison of different NLGM architectures. All structures here are without FFG and ERM.}
\label{tab:NLGMarchitectures}
\end{table}

\subsubsection{Configurations of NLGM:} As shown in Table~\ref{tab:NLGMconfigurations}, we conduct experiments to investigate the performance of different configurations of NLGM. Compared to the baseline (the \nth{1} row in Table~\ref{tab:Ablationanalysis}), the models incorporating SSNLB (the \nth{1} row in Table~\ref{tab:NLGMconfigurations}) or CSNLB (the \nth{2} row in Table~\ref{tab:NLGMconfigurations}) both improve performance on two datasets. The best performance is obtained when using both SSNLB and CSNLB together (the \nth{3} row in Table~\ref{tab:NLGMconfigurations}), showing that these two modules are highly complimentary, and that both spatial and channel-wise non-local features contribute to image saliency.

\section{Conclusion} 
In this paper we seek to address the problem of segmenting multiple salient objects in complex scenes. We present a new architecture for salient object detection, utilising both spatial and channel-wise long-range dependencies. A non-local guidance module captures long-range dependencies between salient objects across the image, allowing the network to better resolve multiple separate salient objects. We also design a feature fusion gate that combines salient and non-local features. The gate utilises progressively refined edge features to promote the most relevant features drawn from each module. Our approach shows state-of-the-art performance across 5 widely used salient object datasets. We also curate an additional dataset comprising only multiple objects in challenging scenes, and show that the performance gap between ours and other methods widens. Consideration of complex scenes is required to further drive development in image saliency, our network and the new dataset can be used as a baseline for performance in this domain.

\bibliography{aaai22}
\end{document}